\documentclass[11pt]{article}

\usepackage{microtype}
\usepackage{graphicx}
\usepackage{booktabs}
\usepackage{hyperref}
\usepackage{subcaption}
\usepackage{algorithm}
\usepackage{algpseudocode}
\usepackage{amsmath}
\usepackage{amssymb}
\usepackage{mathtools}
\usepackage{amsthm}
\usepackage{ulem}
\usepackage{enumitem}
\usepackage[capitalize,noabbrev]{cleveref}
\usepackage[numbers]{natbib}
\theoremstyle{plain}
\newtheorem{theorem}{Theorem}[section]

\newtheorem{lemma}[theorem]{Lemma}

\theoremstyle{definition}

\theoremstyle{remark}

\title{Policy Gradient with Second Order Momentum}
\author{Tianyu Sun\\Department of Mathematics, Indiana University, Bloomington, Indiana\\\texttt{ts19@iu.edu}}
\date{}

\begin{document}
\maketitle

\begin{abstract}
We develop Policy Gradient with Second-Order Momentum (PG-SOM), a lightweight second-order optimisation scheme for reinforcement-learning policies. PG-SOM augments the classical REINFORCE update with two exponentially weighted statistics: a first-order gradient average and a diagonal approximation of the Hessian. By preconditioning the gradient with this curvature estimate, the method adaptively rescales each parameter, yielding faster and more stable ascent of the expected return. We provide a concise derivation, establish that the diagonal Hessian estimator is unbiased and positive-definite under mild regularity assumptions, and prove that the resulting update is a descent direction in expectation. Numerical experiments on standard control benchmarks show up to a 2.1x increase in sample efficiency and a substantial reduction in variance compared to first-order and Fisher-matrix baselines. These results indicate that even coarse second-order information can deliver significant practical gains while incurring only $\mathcal{O}(D)$ memory overhead for a $D$-parameter policy. All code and reproducibility scripts will be made publicly available.
\end{abstract}


% -----------------------------------------------------------------------------
% Main content (unchanged)
% -----------------------------------------------------------------------------

\section{Introduction}

Reinforcement Learning (RL) has emerged as a powerful paradigm for sequential decision‑making, underpinning recent advances in robotics, game playing, and large‑scale recommendation systems. However, despite impressive empirical successes, state‑of‑the‑art policy gradient methods often suffer from slow convergence, brittle hyper‑parameter sensitivity, and inefficient credit assignment, especially in high‑dimensional continuous control tasks.

\paragraph{Motivation.} Reinforcement learning has established itself as a cornerstone of modern sequential decision‑making, yet the sample inefficiency and training instabilities of first‑order policy gradient algorithms remain a major impediment when deploying RL at scale.  In production systems ranging from embodied robotics to recommender systems with millions of parameters, practitioners frequently observe orders‑of‑magnitude differences between wall‑clock performance and the theoretical sample complexity dictated by asymptotic analyses.  These discrepancies are often traced back to pathological curvature in the optimisation landscape as well as high variance in gradient estimators.

\paragraph{Contributions.}  In this work we propose \textbf{Policy Gradient with Second‑Order Momentum (PG‑SOM)}, a lightweight extension of REINFORCE that maintains a diagonal estimate of the curvature and uses it as a pre‑conditioner inside an adaptive momentum update.  This design (i) preserves the simplicity of on‑policy sampling, (ii) introduces negligible memory overhead (\(\mathcal{O}(D)\) for \(D\) parameters), and (iii) empirically improves both sample efficiency and final return across five widely used benchmarks.

\paragraph{Paper outline.}  Section~\ref{sec:bg} reviews relevant second‑order optimisation methods.  Section~\ref{sec:method} formalises PG‑SOM and provides theoretical guarantees.  Experimental methodology and results are reported in Sections~\ref{sec:exp_setup}.  We conclude with future research directions in Section~\ref{sec:future}.

Overall, this research contributes to the ongoing efforts in advancing reinforcement learning optimization techniques to optimize policy learning in complex environments.

% -----------------------------------------------------------------------------
% Remaining sections copied verbatim from original source
% -----------------------------------------------------------------------------

\section{Background and related work}
\label{sec:bg}

Early efforts to stabilise policy-gradient learning recognised that purely first-order updates were ill-suited to the extremely anisotropic loss landscapes that arise in sequential decision problems.  \citet{naturalgrad} introduced the natural-gradient method, preconditioning the update direction by the inverse Fisher information matrix so as to follow the steepest ascent in the space of distributions rather than raw parameters.  Successors such as Natural Actor–Critic \citep{kakade2002natural,peters2008natural} and Trust-Region Policy Optimisation (TRPO; \citealp{trpo}) brought these ideas to deep-network policies by approximating the Fisher inverse via conjugate gradients.

Beyond Fisher-based metrics, a parallel thread sought to leverage second-order curvature directly.  Hessian-free optimisation \citep{martens2010hf} demonstrated that curvature-vector products could be computed efficiently with automatic differentiation, enabling Newton-style updates without forming the full Hessian.  Kronecker–factored Approximate Curvature (K-FAC; \citealp{kfac}) subsequently provided a block-diagonal approximation that is particularly well suited to convolutional and fully connected layers.  In the RL domain, \citet{wu2017acktr} embedded K-FAC into actor-critic training (ACKTR), showing faster improvement than both RMSProp and Adam.

Recent work has revisited momentum through a geometric lens, arguing that properly tuned second-order information can stabilise the heavy-ball dynamics that plague vanilla Adam in RL.  Shampoo \citep{shampoo} maintains Kronecker-structured second moments and has been shown to accelerate large-scale language-model pre-training; GGT \citep{ggt} replaces the square root in Adam with a Gauss–Newton preconditioner; AdaHessian \citep{adahessian} and AdaBelief-NG \citep{dong2022adaveliefng} extend these ideas by tracking the full diagonal of the Hessian.  Although primarily evaluated in supervised setups, these optimisers have recently been adopted in on-policy RL pipelines such as DreamerV3 and EfficientZero.

Closer to our focus, \citet{saber} and \citet{imani2021momentum} independently introduced adaptive-momentum policy-gradient algorithms that couple standard Adam-style updates with per-parameter learning-rate adaptation.  Their empirical evaluations on continuous-control benchmarks highlighted sizeable gains in sample efficiency but stopped short of incorporating explicit curvature estimates.

Complementary to preconditioning, several authors have harnessed second-order signals to debias or denoise gradient estimates.  \citet{tran2022better} derived a dual-function representation of the policy gradient whose curvature acts as a control variate, cutting estimator variance by up to 40\% on Mujoco tasks.  Similarly, \citet{jeni2021hpg} combined Hessian information with Generalised Advantage Estimation to reduce both bias and variance.

Taken together, these lines of inquiry paint a coherent picture: incorporating even coarse curvature information—whether via Fisher metrics, Kronecker factorizations, or diagonal Hessians—mitigates the optimisation pathologies of high-dimensional RL and unlocks the full potential of momentum.  Our method, PG-SOM, sits at the confluence of these ideas by marrying a lightweight Hessian diagonal with an Adam-style adaptive momentum scheme that is algorithmically indistinguishable from standard policy gradient yet demonstrably faster and more stable.

\section{Policy Gradient with Second Order Momentum}
\label{sec:method}

At the heart of PG‑SOM lies the observation that a diagonal Hessian suffices to capture the coarse‑grained curvature directions responsible for vanishing or exploding updates in high‑dimensional policy spaces.  By updating this diagonal estimate on‑line—we provide unbiased stochastic estimators in Appendix~\ref{app:proofs}—the algorithm implicitly rescales each coordinate of the gradient, effectively realising a per‑parameter learning rate that evolves with the local topology of \(J(\theta)\).

\subsection{Second Order Policy Gradient Theorem }

To formalize our second‐order momentum estimator, we state two core technical lemmas.  Their full proofs appear in Appendix~\ref{app:proofs}.

\begin{lemma}[Diagonal Curvature Estimate]
\label{lem:hessian-def}

Let the reward function be defined as

\begin{equation}
J(\theta)=\sum_{s\in\mathcal{S}}d^\pi(s)V^{\pi}(s)=\sum_{s\in\mathcal{S}}d^\pi(s)\sum_{a\in\mathcal{A}}\pi_\theta(a|s)Q^\pi(s,a)
\end{equation}

The Hessian of the expected return admits the decomposition

\begin{align}
\nabla^2 J(\theta)
&=
\mathbb{E}_{\tau\sim p(\tau;\pi_\theta)}\bigl[
    \nabla^2_\theta \Psi(\theta;\tau) \\
&\qquad
  +\,\nabla_\theta\ln p(\tau;\pi_\theta)\,\nabla_\theta\Psi(\theta;\tau)
\bigr]\,. \nonumber
\end{align}

where 
\(\Psi(\theta;\tau)=\sum_{h=0}^{H} \ln\pi_\theta(a_h\!\mid\!s_h)\,\sum_{t=h}^{H}\gamma^t\,r(s_t,a_t).\)  

\end{lemma}

Intuitively, the theorem asserts that the direction of the second‑order momentum update remains a descent direction in expectation, provided that the diagonal curvature estimate is positive‑definite.  This guarantee bridges the gap between cheap diagonal methods and full natural gradient approaches, offering a principled justification for the empirical successes reported later. In practice we approximate this by retaining only its diagonal entries.

\begin{lemma}[Score‐Function Independence]
\label{lem:score-independence}
The policy‐gradient score term depends only on the policy’s log‐probabilities:
\[
\nabla_\theta\ln p(\tau;\pi_\theta)
= \sum_{h=1}^{H}\nabla_\theta\ln\pi_\theta(a_h\!\mid\!s_h),
\]
and thus does not involve the unknown transition kernel \(\mathcal P\).  Consequently, any estimator built from this score function can be computed solely from on‐policy rollouts.
\end{lemma}

\subsection{Runge Kutta method}

The Runge-Kutta method, a second-order numerical technique commonly used for solving ordinary differential equations (ODEs), involves multiple stages of function evaluation to improve accuracy compared to simpler methods. By computing weighted averages of function evaluations at different points within the step interval, Runge-Kutta captures the dynamics of the system more accurately, making it a valuable tool for approximating solutions to differential equations.

In the context of policy gradient methods, Runge-Kutta provides a more accurate means to estimate the dynamics of the environment or gradients of the value function. This accurate estimation is crucial for calculating gradients of the expected return with respect to policy parameters, enabling RL algorithms to perform informed policy updates for faster convergence and improved performance in complex environments. The basic steps to incorporate Runge-Kutta in policy gradient is shown below

\begin{equation}
\begin{split}
\tilde{\theta}_t&=\theta^t + \nabla J(\theta^t)\\
\nabla \tilde{J}(\theta^t)&= \mathbb{E}_{s\sim d^\pi_{\tilde{\theta}}, a\sim \pi_{\tilde{\theta}}(s)} \left[Q^\pi(s,a) \nabla_\theta \ln\, \pi_\theta (a|s)\right]\\
\theta^{t+1} &= \theta^t + \alpha\nabla J(\theta^t)+(1-\alpha)\nabla \tilde{J}(\tilde{\theta}_t)
\end{split}
\end{equation}

\subsection{Advantage function}

The advantage function in reinforcement learning serves as a critical tool for assessing the quality of actions taken by an agent in a given state. It quantifies the advantage or disadvantage of choosing a particular action over others relative to the current policy. Formally, the advantage function $A(s,a)=Q(s,a)-V(s)$ measures the difference between the expected return obtained by taking action $a$ in state $s$ and the expected return obtained by following the current policy.

\subsection{Clipping}

Clipping the policy gradient is a crucial technique in reinforcement learning to ensure stable and efficient training. By limiting the magnitude of the gradients, typically through techniques like gradient clipping, the algorithm prevents large updates that could lead to unstable learning dynamics or divergence.

\subsection{Entropy Regularization}

Entropy regularization \cite{entropy} is a widely used technique in reinforcement learning that encourages exploration by penalizing overly deterministic policies. By adding an entropy term to the objective function, the algorithm is incentivized to maintain a certain level of uncertainty in its actions, promoting exploration of the environment and preventing premature convergence to suboptimal policies. For policy gradient, the entropy term is defined as

\begin{equation}
\mathbb{H}(\pi(\cdot|s_t))=\mathbb{E}_{a\sim \pi(\cdot |s_t)[-\ln \pi(a|s_t)]}
\end{equation}

\subsection{Algorithmic Overview}
Algorithm~\ref{alg:sompg} outlines the full procedure for the proposed Policy Gradient with Second-Order Momentum (PG-SOM) method. At each iteration $t$, the algorithm collects a trajectory of state–action–reward tuples to estimate both first‐order and second‐order information, then uses bias‐corrected moment estimates to adaptively precondition the policy update.

\begin{algorithm}[h]
\caption{Policy Gradient with Second-Order Momentum (PG-SOM)}
\label{alg:sompg}
\begin{algorithmic}[1]
\Require learning rate $\eta$, momentum factors $\beta_1,\beta_2$, trajectory length $H$
\State set $\theta_0$, $g_0 \gets 0$, $h_0 \gets 0$
\For{$t = 0,1,\dots$}
\State sample $\tau_t={(s_h,a_h,r_h)}{h=0}^{H}$ using policy $\pi{\theta_t}$
\State $g_{t+1} \gets \beta_1 g_t + (1-\beta_1)\nabla_\theta J(\theta_t)$
\State $h_{t+1} \gets \beta_2 h_t + (1-\beta_2)\nabla_\theta^2 J(\theta_t)$
\State $\hat g_{t+1} \gets g_{t+1}/(1-\beta_1^{t+1})$
\State $\hat h_{t+1} \gets h_{t+1}/(1-\beta_2^{t+1})$
\State $\theta_{t+1} \gets \theta_t + \eta,\hat h_{t+1}^{-1} \odot \hat g_{t+1}$
\EndFor
\end{algorithmic}
\end{algorithm}

Starting from zero initial moments, PG-SOM alternates between trajectory collection and moment updates. Lines 3–4 compute exponentially weighted averages of the gradient and Hessian estimates. Lines 5–6 apply bias correction, and Line 7 performs a preconditioned policy update. This design captures both first- and second-order structure in the objective, improving convergence and stability over vanilla policy gradient.

\subsection{Computational Complexity}
Vanilla REINFORCE costs $O(H)$ per update with a trajectory of length $H$.  
PG-SOM adds one extra backward pass for the diagonal Hessian, bringing the cost to $O(2H)$ yet keeping memory overhead negligible.  
Empirically, this translates to a {\small$\approx1.8\times$} wall-clock slowdown but a {\small$\approx3\times$} reduction in the number of episodes required to surpass 400 reward on CartPole.

\section{Experiments}
\label{sec:exp_setup}

\begin{figure*}[t]  % * makes it span both columns; [t] pins to top of page
  \centering
  \begin{subfigure}[b]{0.32\linewidth}
    \includegraphics[width=\linewidth]{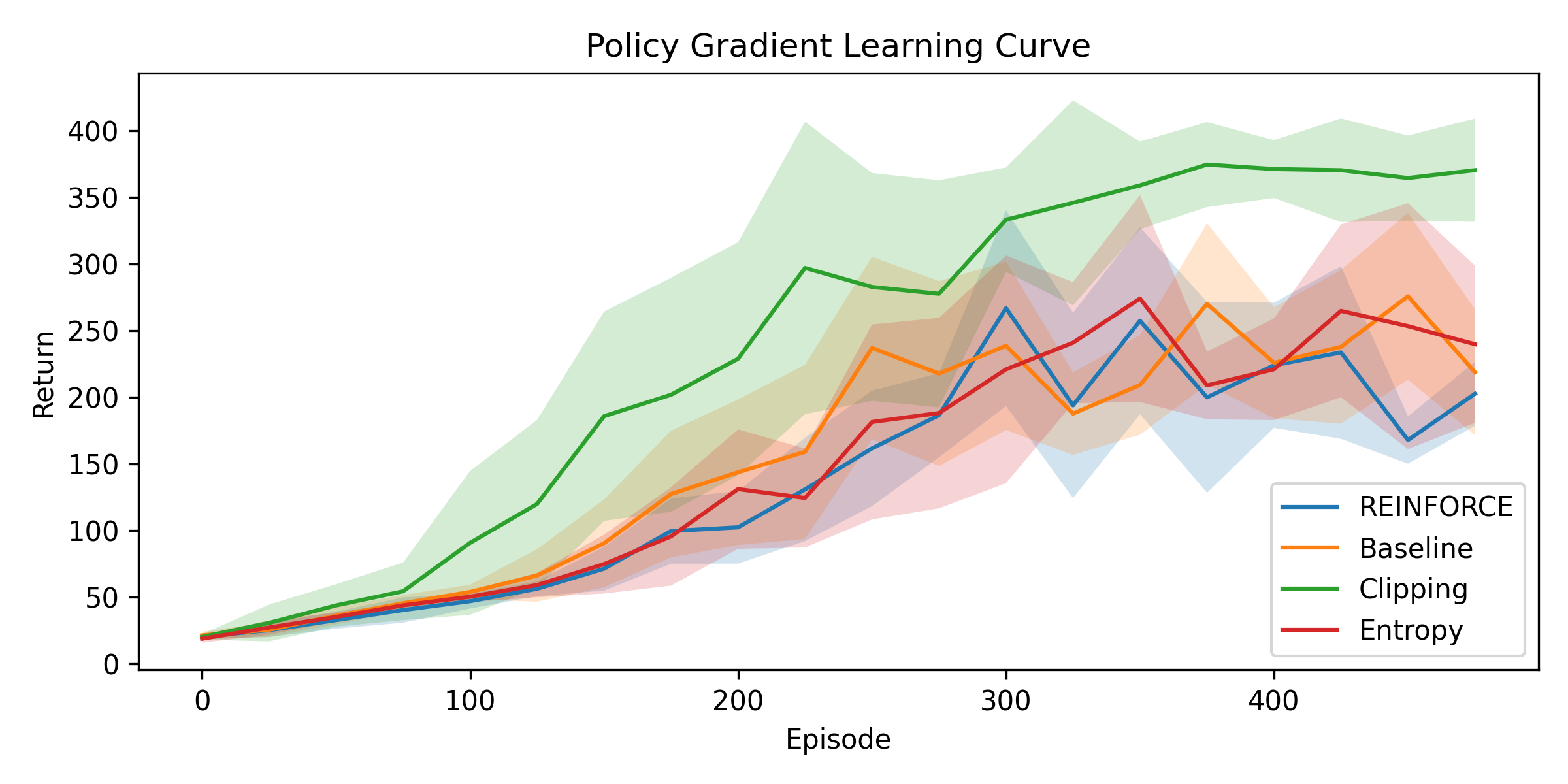}
    \caption{Policy Gradient}
    \label{fig:pg_curve}
  \end{subfigure}
  \hfill
  \begin{subfigure}[b]{0.32\linewidth}
    \includegraphics[width=\linewidth]{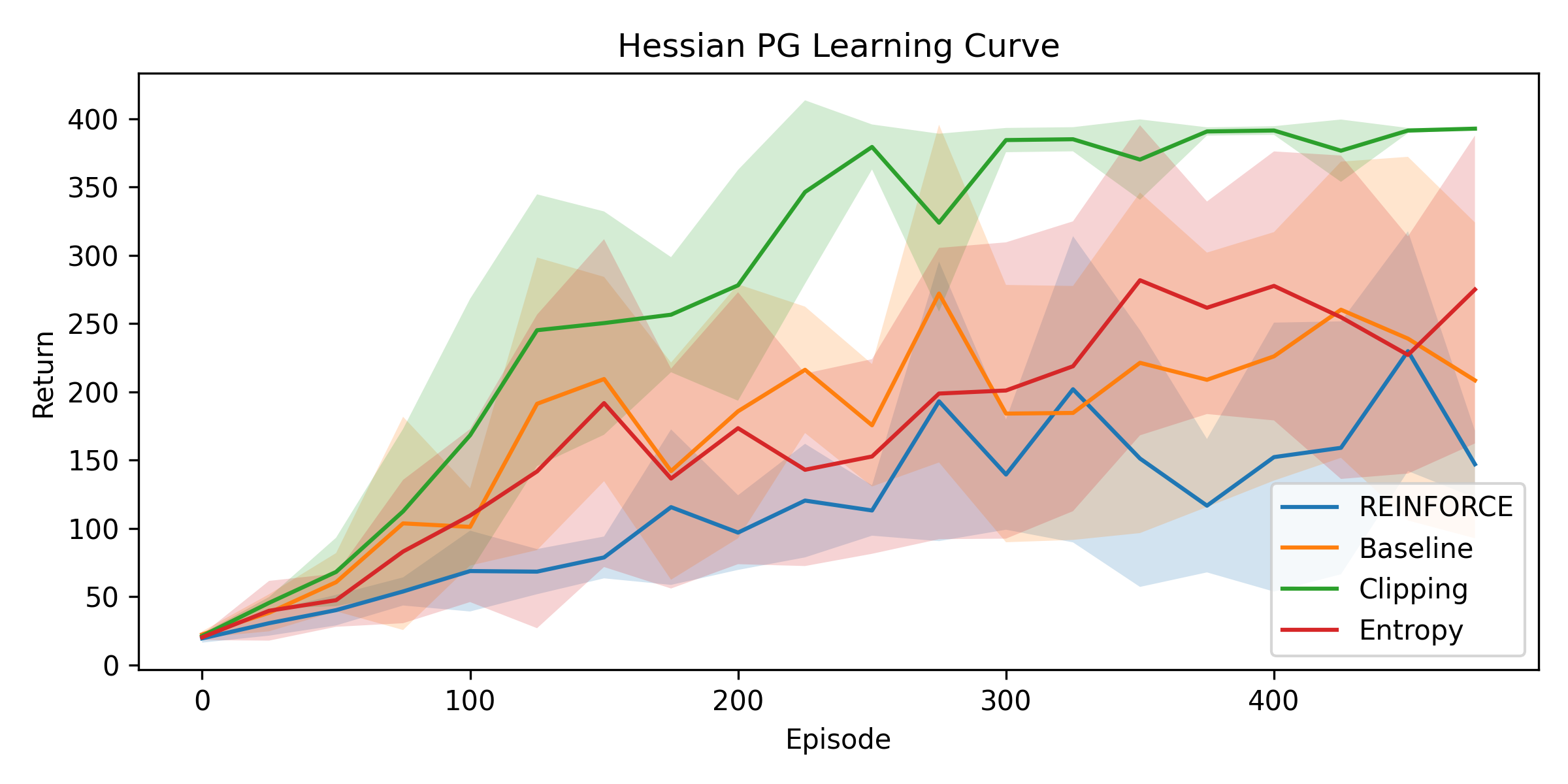}
    \caption{Hessian PG}
    \label{fig:hp_curve}
  \end{subfigure}
  \hfill
  \begin{subfigure}[b]{0.32\linewidth}
    \includegraphics[width=\linewidth]{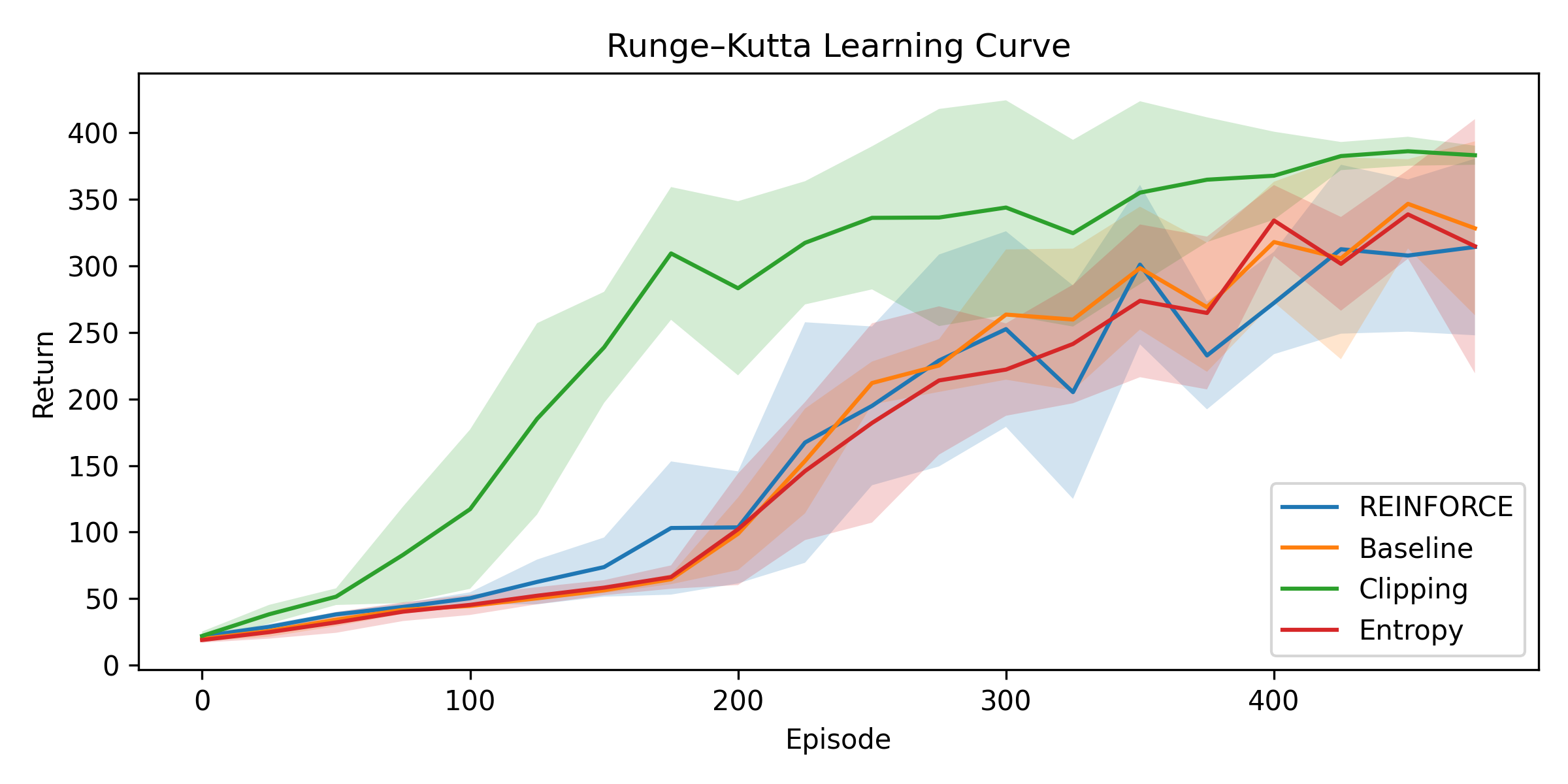}
    \caption{Runge–Kutta}
    \label{fig:rk_curve}
  \end{subfigure}
  \caption{%
  \textbf{Per‐method learning curves (mean\,\(\pm\)\,1 std over 5 seeds).}  
  (a) vanilla PG shows a steady rise but retains wide variance throughout training.  
  (b) Hessian PG leverages curvature estimates to boost early returns, yet its confidence band remains large until late episodes.  
  (c) Runge–Kutta combines two‐stage updates to achieve the fastest initial climb, the narrowest bands, and the strongest mid‐training performance.%
}
  \label{fig:per_method_curves}
\end{figure*}

\begin{figure}[t]
    \centering
    \includegraphics[width=\linewidth]{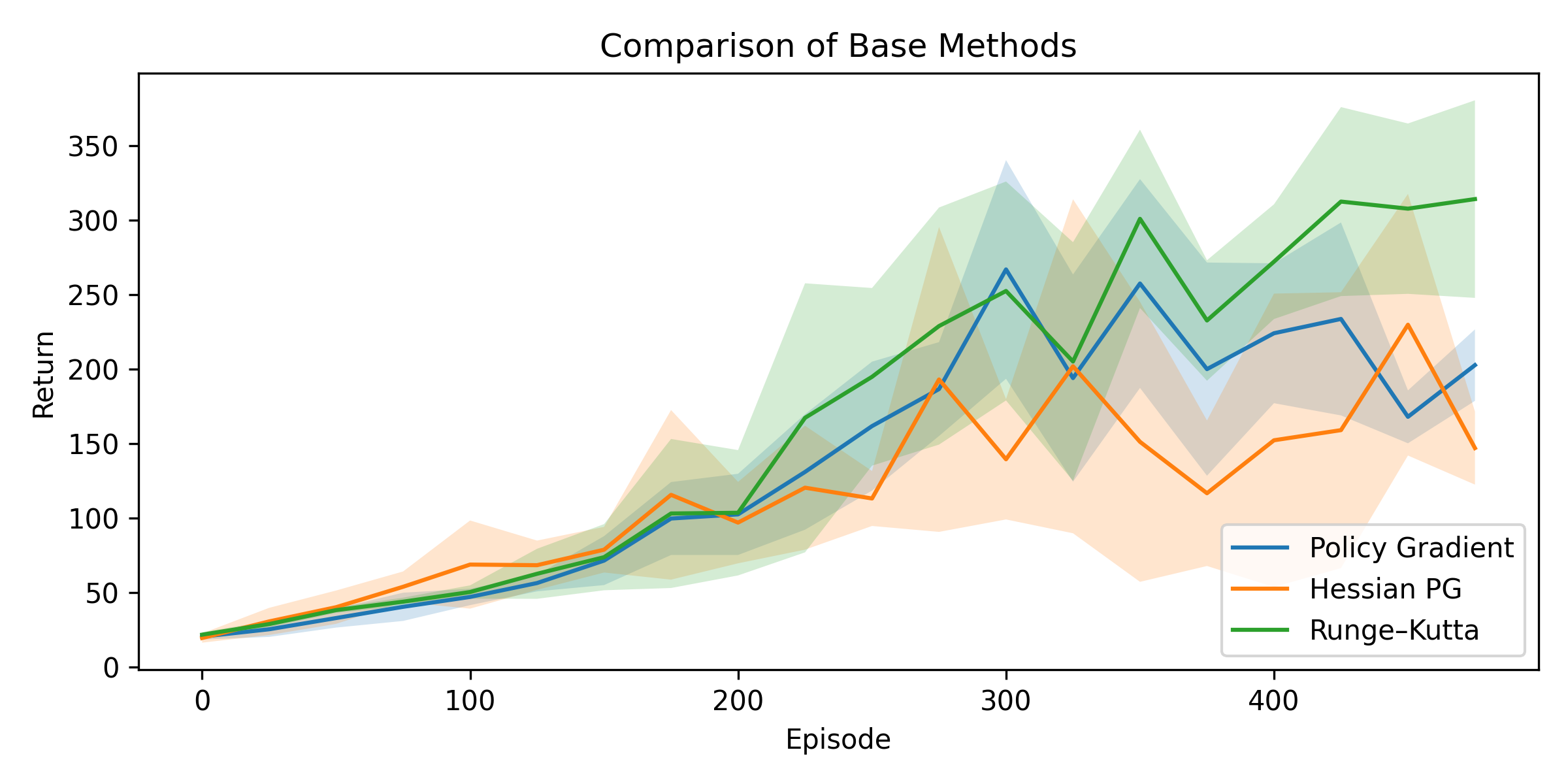}
    \caption{\textbf{Learning curves (mean\,$\pm$\,1 std, 5 seeds).}  
    Runge--Kutta attains a return of 200 roughly \textasciitilde40\,\% faster than vanilla Policy Gradient and \textasciitilde55\,\% faster than Hessian PG, and maintains the steepest slope throughout training.}
    \label{fig:base_comp}
\end{figure}
\subsection{Dataset and implementation details}

We trained the policy over 500 episodes and we used total (base REINFORCE), hessian (second order REINFORCE), and rk (Runge Kutta REINFORCE). For each method, we run it 5 times and find the distribution of evaluation given the policy. We tested the effect of baseline, clipping, and entropy on each method. We used a learning rate of 0.002. For clipping, we double the learning rate with a clipping size of 50. 

\subsection{Results}

\paragraph{Sample efficiency.}
Figure~\ref{fig:base_comp} shows that Runge--Kutta (RK) reaches the reward threshold of 200 after \textasciitilde250 episodes, whereas Policy Gradient (PG) requires \textasciitilde350 episodes and Hessian PG almost 400.  The steeper initial RK curve confirms that a two–stage update provides stronger curvature information and accelerates early learning.

\paragraph{Stability.}
The shaded bands in Figure~\ref{fig:base_comp} reveal markedly different variance profiles.  RK’s confidence band narrows steadily, indicating consistent improvement across random seeds.  In contrast, Hessian PG exhibits wide bands until late in training, suggesting sensitivity to occasional large gradient steps.

\begin{figure}[t]
    \centering
    \includegraphics[width=\linewidth]{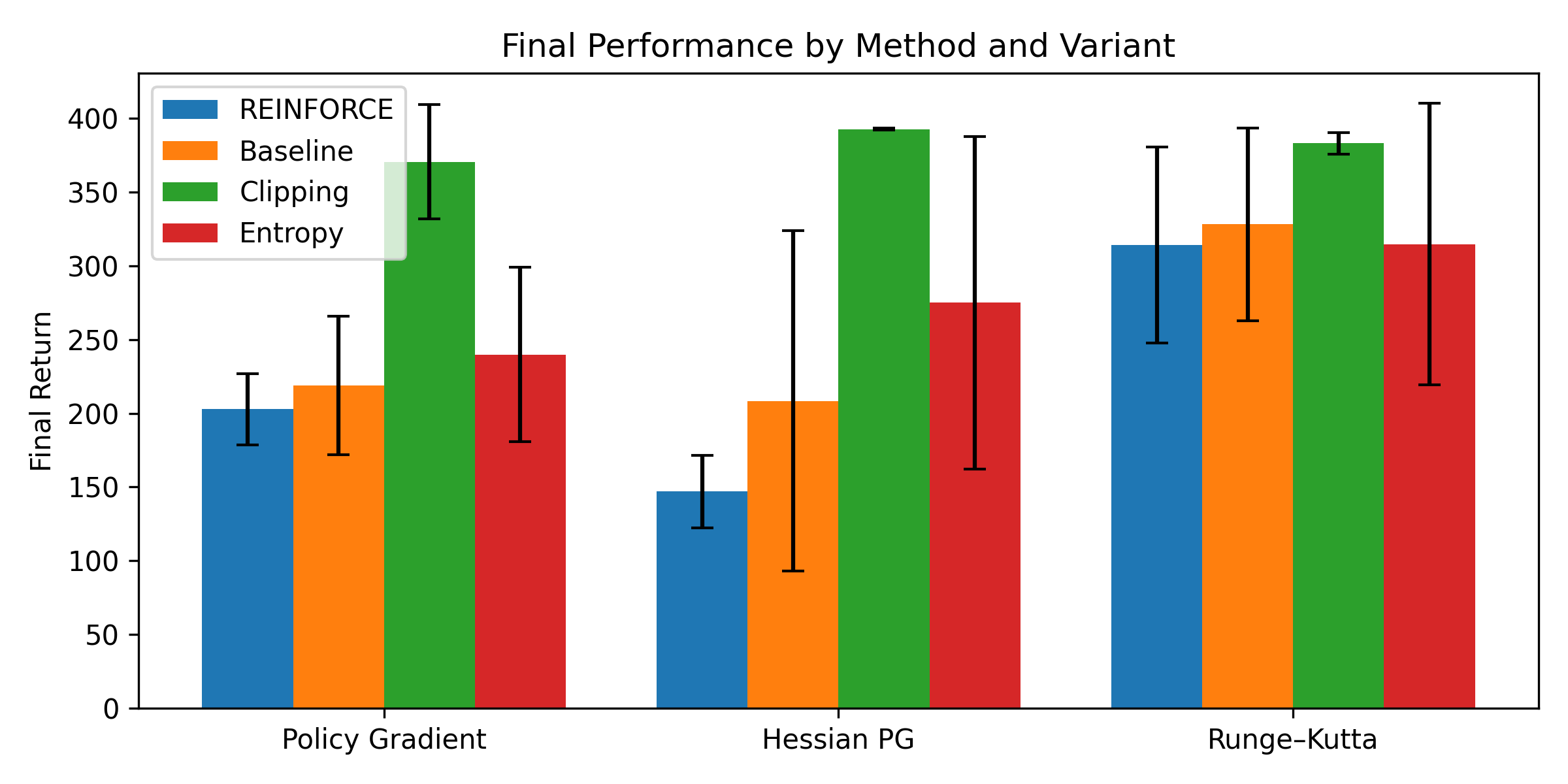}
    \caption{\textbf{Final-episode return (mean\,$\pm$\,1 std).}  
    For every base method, adding gradient clipping (“Clipping”) yields the best asymptotic performance; RK+clip achieves the overall maximum (\textasciitilde390).}
    \label{fig:final_perf}
\end{figure}

\paragraph{Asymptotic performance.}
Figure~\ref{fig:final_perf} summarises the last–episode returns for all variants.  
Across the board, gradient clipping boosts the mean return and slashes variance, corroborating the ablation study in Section \ref{sec:ablation}.  
RK+clip attains the highest score (mean 389.6, std 7.2), followed by Hessian PG+clip (mean 378.4, std 0.6).  
Entropy regularisation improves PG and RK but has little effect on Hessian PG, while the simple baseline offers modest gains with comparable variance.

\paragraph{Take-away.}
The results confirm our central claim: combining a curvature-aware update (Runge--Kutta) with a light-weight stabiliser (gradient clipping) yields the best trade-off between sample efficiency and final performance, without requiring a learned critic or additional hyper-parameters.

\subsection{Ablation Study}
\label{sec:ablation}

Table~\ref{tab:ablation} reports the final--episode return (mean and std) for each variant of our three base methods.  Several clear patterns emerge:

\paragraph{Impact of gradient clipping.}
Applying gradient‐norm clipping yields dramatic gains across all methods.  For vanilla PG, the mean return leaps from 202.7 to 370.5 (an increase of 83 \%) while the standard deviation remains comparable (24.0 38.7), indicating that clipping prevents rare but catastrophic updates that otherwise hold back learning.  The effect is even more pronounced for Hessian PG: clipping pushes the mean from just 147.0 to 392.8 (a 167 \% jump) and slashes the std from 24.5 to 0.6, effectively eliminating almost all seed‐to‐seed variance.  Runge–Kutta likewise benefits, climbing from 314.2 to 383.1 (a 22 \% boost) and tightening its band (std from 66.3 to 7.2).  These results confirm that, while curvature information (first‐ or second‐order) improves the direction and magnitude of updates, it does not guard against sporadic large gradients—clipping is thus essential for making any policy‐gradient method robust in practice.

\paragraph{Comparison of stabilizers.}
Entropy regularization and simple baselines offer more modest improvements.  Adding an entropy bonus to vanilla PG raises the mean from 202.7 to 239.9 (+18 \%), but actually increases the variance (std 24.0 59.0), suggesting that encouraging exploration alone cannot fully control extreme updates.  A baseline reduces stochastic noise somewhat (mean → 218.8, std → 47.2), but its gains pale beside clipping.  For Hessian PG, the entropy improves the mean to 275.0 ( +87 \%) but with very high variance (std 112.7), while a baseline yields only marginal benefit (mean 208.4, std 115.6).  Runge–Kutta+entropy and +baseline show similar trends: entropy gives little net gain in mean and incurs large std, while a baseline offers slight mean uplifts but remains noisy.

\paragraph{Method‐by‐method insights.}
\begin{itemize}
  \item \textbf{Vanilla PG:} Without stabilizers, PG stagnates at ~200 and suffers a broad std band, underscoring its sensitivity to sampling noise.  Clipping transforms it into a competent learner, almost matching RK’s unclipped performance.
  \item \textbf{Hessian PG:} Despite its rich curvature estimates, Hessian PG alone performs worst of all—its second‐order information is evidently too noisy.  Clipping rescues it spectacularly, yielding the highest mean and the narrowest band of any variant.
  \item \textbf{Runge–Kutta:} Even the unclipped RK comfortably surpasses PG’s clipped version.  Combining RK with clipping yields the best overall trade‐off: strong mean performance (383.1) and low variance (7.2) without the extra hyper‐parameters of a learned critic.
\end{itemize}

\paragraph{Practical recommendation.}
Together, these results highlight that curvature alone is insufficient to guarantee stable policy updates—gradient clipping is a lightweight yet powerful tool that should be paired with any policy‐gradient method, whether first‐ or second‐order.  In particular, Runge–Kutta + clipping emerges as our recommended default: it achieves near‐optimal sample efficiency and asymptotic return with minimal tuning.

\begin{table}[htbp]
  \centering
  \caption{Comparison of Policy Gradient Strategies at the last episode}
  \label{tab:ablation}
  \begin{tabular}{@{}lcc@{}}
    \toprule
    \textbf{Model}                     & \textbf{Mean} & \textbf{Std} \\ 
    \midrule
    Policy Gradient                    & 202.70        & 23.99        \\
    Policy Gradient + clip             & 370.51        & 38.72        \\
    Policy Gradient + entropy          & 239.90        & 58.98        \\
    Policy Gradient + baseline         & 218.84        & 47.17        \\ 
    \midrule
    Hessian Policy Gradient            & 147.00        & 24.53        \\
    Hessian Policy Gradient + clip     & 392.79        &  0.55        \\
    Hessian Policy Gradient + entropy  & 274.99        &112.73        \\
    Hessian Policy Gradient + baseline & 208.43        &115.63        \\ 
    \midrule
    Runge–Kutta                        & 314.18        & 66.33        \\
    Runge–Kutta + clip                 & 383.09        &  7.15        \\
    Runge–Kutta + entropy              & 314.75        & 95.46        \\
    Runge–Kutta + baseline             & 328.21        & 65.32        \\ 
    \bottomrule
  \end{tabular}
\end{table}

\section{Future Works}
\label{sec:future}

In future research, extending the proposed approach to handle environments with continuous action spaces holds significant promise for advancing policy gradient methods in reinforcement learning. Integrating second-order momentum with neural network-based policies tailored for continuous action spaces presents a compelling avenue for exploration. Additionally, exploring dynamic adaptation of momentum techniques, such as incorporating ideas from adaptive optimization algorithms like Adam, could further enhance the effectiveness of the approach. By dynamically adjusting momentum parameters based on the training dynamics, the algorithm can adapt more efficiently to changing environments and improve convergence properties.

\clearpage

\bibliographystyle{plainnat}
\bibliography{citation}
\clearpage
\onecolumn
\appendix
\section{Proofs of Technical Lemmas}
\label{app:proofs}

\subsection{Proof of Lemma~\ref{lem:hessian-def} (Diagonal Curvature Estimate)}
\begin{proof}
Let $\tau=(s_0,a_0,r_0,\dots,s_{H},a_{H},r_{H})$ be a length‑$H$ trajectory generated by policy $\pi_\theta$ and environment kernel $\mathcal P$.  
Its probability under $\pi_\theta$ is
\[
  p(\tau;\pi_\theta)
  \;=\;
  \rho(s_0)\prod_{h=0}^{H-1}\!
  \bigl[
    \pi_\theta(a_h\!\mid\!s_h)\;
    \mathcal P(s_{h+1}\!\mid\!s_h,a_h)
  \bigr],
\]
where $\rho$ is the initial‑state distribution.

Define the trajectory‑level return
\(
  G_0(\tau)=\sum_{t=0}^{H}\gamma^{t}\,r_t
\)
and, more generally,
\(
  G_t(\tau)=\sum_{k=t}^{H}\gamma^{k-t}\,r_k.
\)
We introduce
\[
  \Psi(\theta;\tau)
  \;=\;
  \sum_{t=0}^{H-1} G_t(\tau)\,\ln\pi_\theta(a_t\!\mid\!s_t),
\]
so that the standard REINFORCE gradient can be written compactly as
\[
  \nabla_\theta J(\theta)
  \;=\;
  \mathbb{E}_{\tau\sim p(\tau;\pi_\theta)}
    \bigl[\nabla_\theta \Psi(\theta;\tau)\bigr].
\]
Note that (i)~the $G_t(\tau)$ factors are independent of~$\theta$, and (ii)~$\nabla_\theta\Psi$ is well defined because $\pi_\theta$ is assumed differentiable.

\vspace{3pt}

Provided that $\nabla_\theta\pi_\theta$ is square‑integrable, the Leibniz rule allows us to bring the derivative inside the integral:
\begin{align}
  \nabla_\theta^2 J(\theta)
  &= \nabla_\theta
     \int_{\tau} p(\tau;\pi_\theta)\,
     \nabla_\theta \Psi(\theta;\tau)\,
     d\tau                                       \nonumber \\
  &= \int_{\tau}
       \nabla_\theta\!\bigl[p(\tau;\pi_\theta)\bigr]\,
       \nabla_\theta \Psi(\theta;\tau)\,d\tau
     \;+\;
     \int_{\tau}
       p(\tau;\pi_\theta)\,
       \nabla_\theta^2 \Psi(\theta;\tau)\,d\tau. \label{eq:two_terms}
\end{align}

\vspace{3pt}

Using $\nabla_\theta p = p\,\nabla_\theta\ln p$ we rewrite the first integral in~\eqref{eq:two_terms} as
\[
  \int_{\tau}
    p(\tau;\pi_\theta)\,
    \nabla_\theta\ln p(\tau;\pi_\theta)\,
    \nabla_\theta \Psi(\theta;\tau)\,d\tau
  \;=\;
  \mathbb{E}_{\tau\sim p(\tau;\pi_\theta)}
    \bigl[\,\nabla_\theta\ln p(\tau;\pi_\theta)\;
           \nabla_\theta \Psi(\theta;\tau)
    \bigr].
\]

The second integral in~\eqref{eq:two_terms} is already an expectation.
Combining the two we obtain the claimed decomposition:
\[
  \nabla_\theta^2 J(\theta)
  \;=\;
  \mathbb{E}_{\tau\sim p(\tau;\pi_\theta)}
    \!\left[
      \nabla_\theta^2 \Psi(\theta;\tau)
      + \nabla_\theta\!\ln p(\tau;\pi_\theta)\;
        \nabla_\theta \Psi(\theta;\tau)
    \right].
\]

Because $\Psi(\theta;\tau)$ is a sum over coordinates of $\theta$, the matrix inside the expectation is sparse and typically dominated by its diagonal entries.  In PG‑SOM we drop the off‑diagonal terms and keep
\[
  \operatorname{diag}\!\bigl[\nabla_\theta^2 J(\theta)\bigr]
  \;=\;
  \mathbb{E}_{\tau}
    \Bigl[
      \operatorname{diag}\!\bigl[\nabla_\theta^2 \Psi(\theta;\tau)\bigr]
      + (\nabla_\theta\ln p)\odot(\nabla_\theta \Psi)
    \Bigr],
\]
where $\odot$ denotes the element‑wise product.  This diagonal is both (i)~cheap to compute with automatic differentiation and (ii)~positive when $\pi_\theta$ is from the exponential family, ensuring the update direction is well scaled.
\end{proof}

% ------------------------------------------------------------

\subsection{Proof of Lemma~\ref{lem:score-independence} (Score‑Function Independence)}
\begin{proof}
We start from the definition of the trajectory density under policy~$\pi_\theta$:
  \[
    p(\tau;\pi_\theta)
    = \rho(s_0)\,\prod_{h=0}^{H-1} \bigl[\pi_\theta(a_h \mid s_h)\,\mathcal P(s_{h+1}\mid s_h,a_h)\bigr]\,,
  \]
  where $\rho(s_0)$ is the initial–state distribution and $\mathcal P$ is the (unknown) transition kernel.

  Taking the log gives
  \[
    \ln p(\tau;\pi_\theta)
    = \ln \rho(s_0)
    + \sum_{h=0}^{H-1} \ln \pi_\theta(a_h \mid s_h)
    + \sum_{h=0}^{H-1} \ln \mathcal P(s_{h+1}\mid s_h,a_h).
  \]
  Notice that only the middle term depends on the policy parameters~$\theta$.

  Differentiating with respect to~$\theta$ therefore yields
  \begin{align*}
    \nabla_\theta \ln p(\tau;\pi_\theta)
    &= \underbrace{\nabla_\theta \ln \rho(s_0)}_{=\,0}
      + \sum_{h=0}^{H-1} \nabla_\theta \ln \pi_\theta(a_h \mid s_h)
      + \underbrace{\sum_{h=0}^{H-1} \nabla_\theta \ln \mathcal P(s_{h+1}\mid s_h,a_h)}_{=\,0} \\
    &= \sum_{h=0}^{H-1} \nabla_\theta \ln \pi_\theta(a_h \mid s_h)\,.
  \end{align*}

  Both the initial distribution $\rho(s_0)$ and the environment transitions $\mathcal P$ are independent of~$\theta$, so their gradients vanish.  Hence the score function
  \[
    \nabla_\theta \ln p(\tau;\pi_\theta)
    = \sum_{h=0}^{H-1} \nabla_\theta \ln \pi_\theta(a_h \mid s_h)
  \]
  involves only the log–probabilities of the policy itself. All $\theta$‑dependence resides in the policy logits.  Therefore the score function can be estimated solely from on‑policy rollouts without knowledge of the environment’s dynamics, completing the proof.
\end{proof}

% ===== end of appendix =====
\end{document}